\def\year{2020}\relax
\documentclass[letterpaper]{article} 
\usepackage{aaai20}  
\usepackage{times}  
\usepackage{helvet} 
\usepackage{courier}  
\usepackage[hyphens]{url}  
\usepackage{graphicx} 
\usepackage{graphicx}  
\usepackage{amssymb}
\usepackage{amsmath}
\usepackage{subfigure}
\usepackage{multirow}
\usepackage{algorithm}
\usepackage{algorithmicx}
\usepackage{algpseudocode}
\usepackage{diagbox}
\usepackage{color}

\def\ie{{\em i.e.}}
\def\eg{{\em e.g.}}
\def\etal{{\em et al. }}

\title{Relational Learning for Joint Head and Human Detection}
\author{Cheng Chi$^{1,3*}$, Shifeng Zhang$^{2,3}$\thanks{These authors contributed equally to this work.}, Junliang Xing$^{2,3}$, Zhen Lei$^{2,3}$, Stan Z. Li$^{2,3}$, Xudong Zou$^{1,3}$\\
$^{1}$ Institute of Electronics, Chinese Academy of Sciences, Beijing, China\\
$^{2}$CBSR \& NLPR, Institute of Automation, Chinese Academy of Sciences, Beijing, China\\
$^{3}$University of Chinese Academy of Sciences, Beijing, China
}
 
\begin{document}

\maketitle

\begin{abstract}
Head and human detection have been rapidly improved with the development of deep convolutional neural networks. However, these two tasks are often studied separately without considering their inherent correlation, leading to that 1) head detection is often trapped in more false positives, and 2) the performance of human detector frequently drops dramatically in crowd scenes. To handle these two issues, we present a novel joint head and human detection network, namely JointDet, which effectively detects head and human body simultaneously. Moreover, we design a head-body relationship discriminating module to perform relational learning between heads and human bodies, and leverage this learned relationship to regain the suppressed human detections and reduce head false positives. To verify the effectiveness of the proposed method, we annotate head bounding boxes of the CityPersons and Caltech-USA datasets, and conduct extensive experiments on the CrowdHuman, CityPersons and Caltech-USA datasets. As a consequence, the proposed JointDet detector achieves state-of-the-art performance on these three benchmarks. To facilitate further studies on the head and human detection problem, all new annotations, source codes and trained models will be public.
\end{abstract}

\section{Introduction}
\label{sec:intro}
Head and human detection are two important research topics in computer vision field with various applications, such as human behavior analysis, intelligent video surveillance and automatic driving. Although great progress has been made by deep convolutional neural networks (CNNs) on general object detection \cite{DBLP:journals/pami/RenHG017,DBLP:conf/nips/DaiLHS16,DBLP:conf/cvpr/LinDGHHB17,DBLP:conf/eccv/LiuAESRFB16,DBLP:conf/iccv/LinPRK17,DBLP:conf/cvpr/RedmonDGF16}, research in the realm of these two subtasks remains challenging due to their characteristics.

\begin{figure}[t]
\centering
\subfigure[Remove head false positives]{
\label{fig:rm} 
\includegraphics[width=0.525\linewidth]{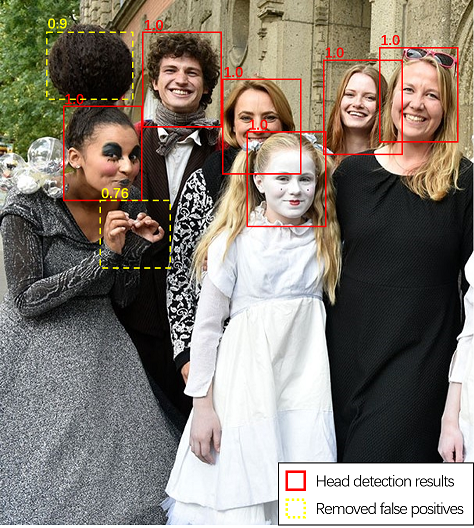}}
\subfigure[Recall suppressed bodies]{
\label{fig:rc} 
\includegraphics[width=0.435\linewidth]{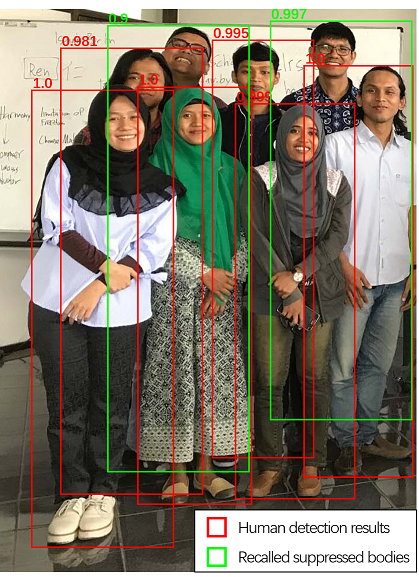}}
\vspace{-3.0mm}
\caption{Effectiveness of JointDet. (a) Remove head false positives: red bounding boxes are the head detection results, and yellow dotted bounding boxes are the removed false positives. (b) Recall missing bodies: red bounding boxes are the human detection results after NMS, and green bounding boxes are the recalled results from suppressed detections.}
\label{fig:effect-post}
\vspace{-3.0mm}
\end{figure}

Head detection has experienced tremendous development in recent years. The context-aware CNN model \cite{DBLP:conf/iccv/VuOL15} employs a pairwise CNN to model pairwise relations among heads. The HeadNet \cite{DBLP:conf/fgr/ChenCHSC18} utilizes spatial semantic relations between pedestrian head and other body parts. However, how to reduce the false positives, such as hair, hands and elbows shown in Figure \ref{fig:rm}, still remains an active research direction. Tracing the main cause, the lack of adequate features and contextual informations is the main difficulty.

As for human detection, occlusion is one of the main challenges, especially in the crowded scenes. Some efforts have been made to handle occlusion. The repulsion loss proposed in \cite{DBLP:conf/cvpr/WangXJSSS18} pushes each proposal not only to approach its designated target, but also to keep it away from the other ground truth objects and their corresponding designated proposals. The attention model \cite{DBLP:conf/cvpr/Zhang0S18} employs an attention network employing an attention mechanism across channels with guidances. The Bi-box model \cite{DBLP:conf/eccv/ZhouY18} proposes a network to simultaneously detect pedestrian  and estimate occlusion by regressing two bounding boxes for full body and visible part estimation respectively. These methods can alleviate the occlusion issue to some extent. However, while it comes to extremely crowded scenes where the overlaps between humans become large, the Non-Maximum Suppression (NMS) post-process method will result in missing a very large portion of targets, as shown in Figure \ref{fig:rc}.

\begin{figure*}[t!]
\centering
\includegraphics[width=1.0\linewidth]{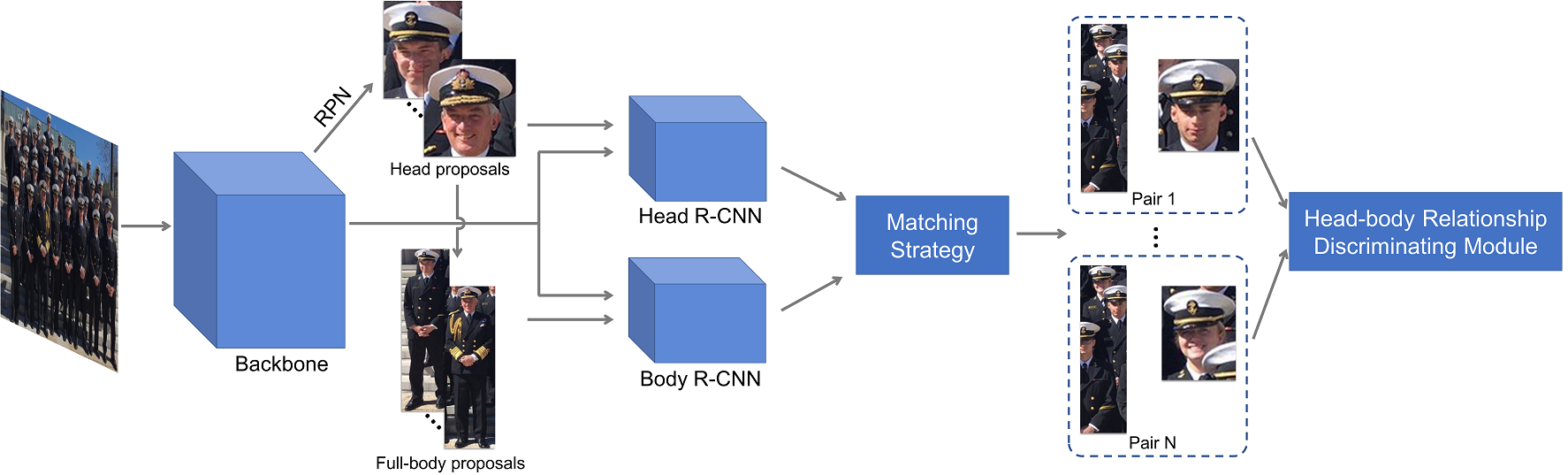}
\caption{Network structure of JointDet. It consists of RPN, Head R-CNN, Body R-CNN and RDM. RPN only generates head proposals, then a statistical head-body ratio is applied to obtain full-body proposals. After that, head and full-body proposals are sent into two parallel R-CNN branches to obtain temporary results. These temporary results are further processed to get final results as follows: (1) matching them using the proposed strategy to output the matched body-head pairs as Pair 1 to Pair N; (2) extracting corresponding features of each pair for RDM to discriminate their relation (\ie, whether they belong to the same person); (3) according to the learned relationship to reduce head false positives and recall suppressed human detections.}
\label{fig:framework}
\end{figure*}

In this paper, we propose a novel joint head and human detection network, namely JointDet, which detects head and human body simultaneously and performs relational learning between them to improve the performance of both two tasks. As shown in Figure~\ref{fig:framework}, we tile a small quantity of anchors with only one scale and one aspect ratio in each pyramid level to generate head proposals in RPN and then a statistical ratio is applied on head proposals to obtain human body proposals, which significantly accelerates both training and inference. These two classes of proposals are sent into two parallel R-CNN to perform second-stage detection. Moreover, we design a head-body Relationship Discriminating Module (RDM) to predict the relationship between heads and bodies. Since even in extremely crowded scenes, the occlusion between heads is not very serious, we utilize the head location to regain the suppressed body detections. On the other hand, due to the lack of adequate features, head detection usually has false positives on elbows, hands and knees. The proposed post-process strategy also reduces these head false positives via the learned relationship.

As mentioned above, both head and human body annotations are necessary for the proposed method, and only the CrowdHuman dataset \cite{DBLP:journals/corr/abs-1805-00123} is publicly available to conduct experiments. To further verify the effectiveness of the proposed model, we annotate head bounding boxes of CityPersons \cite{DBLP:conf/cvpr/ZhangBS17} and Caltech-USA \cite{DBLP:conf/cvpr/DollarWSP09} based on their human body annotations. However, the commonly used $10\times$ training annotations \cite{DBLP:conf/cvpr/ZhangBOHS16} of Caltech-USA are refined automatically with relatively poor quality, it is hard to annotate head bounding boxes based on the original annotations. Therefore, we also re-annotate Caltech-USA with the full-body bounding-box and the visible-region bounding-box, which serves as a satisfied version of Caltech-USA. All these new annotations will be released.

To summarize, this work has five main contributions: 1) proposing an effective framework for joint detection of head and human; 2) designing a RDM to perform relational learning between head and human; 3) introducing a post-process strategy to recall suppressed human detections and reduce head false positives simultaneously; 4) providing the re-annotated body annotations of Caltech-USA, and the head annotations of CityPersons and Caltech-USA to facilitate further studies on the joint detection of head and human; 5) achieving state-of-the-art performance on CrowdHuman, CityPersons and Caltech-USA.

\section{Related Work}

{\flushleft \textbf{Generic Object Detection.}}
Early generic object detectors \cite{DBLP:journals/pami/FelzenszwalbGMR10} rely on the sliding window paradigm based on hand-crafted features and classifiers to find objects of interest. In recent years, a new generation of more effective object detectors based on deep convolutional neural network (CNN) significantly improve the state-of-the-art performances, which can be roughly divided into two categories, \ie, the one-stage approach and the two-stage approach. The one-stage approach \cite{DBLP:conf/eccv/LiuAESRFB16,DBLP:conf/cvpr/RedmonDGF16} directly predicts object class label and regresses object bounding box based on the pre-tiled anchor boxes using deep CNNs. The main advantage of the one-stage approach is its high computational efficiency. In contrast to the one-stage approach, the two-stage approach \cite{DBLP:journals/pami/RenHG017,DBLP:conf/nips/DaiLHS16,DBLP:conf/cvpr/LinDGHHB17} always achieves top accuracy on several benchmarks, which first generates a pool of object proposals by a separated proposal generator (\eg, RPN \cite{DBLP:journals/pami/RenHG017}), and then predicts the class label, accurate location and size of each proposal.

\noindent \textbf{Head Detection.}
Early head detectors are used for crowd counting. Merad \etal \cite{DBLP:conf/avss/MeradAT10} combine positive points of all previous techniques in the head detector. Venkatesh \etal \cite{DBLP:conf/avss/VenkateshDC12} train a head detector using a state-of-the-art cascade of boosted integral features. However, their performance is severely affected under high scene and scale variations because of the usage of handcrafted features. With the arrival of deep learning, some CNN-based methods are proposed. Stewart \etal \cite{DBLP:conf/cvpr/StewartAN16} introduce a proposal-free head detector that is produced from CNN encoders using a regression module, where the regression is generally composed of LSTM so that the variable length output prediction is possible. 
Le \etal \cite{DBLP:conf/icip/LeMWL18} introduce a pairwise head detector based on key parts context of the human head and shoulder, and assisted by priority of scene geometry structure. Vu \etal \cite{DBLP:conf/iccv/VuOL15} predict the scales and the positions of the head directly from the image, then model the pairwise relationships among the objects. Recently introduction of context information is attractive to improve performance. 
Some methods \cite{DBLP:conf/avss/ChenBNT16} exploit depth information for head detection with depth images. Nghiem \etal \cite{DBLP:conf/isspa/NghiemAM12} conduct head detection on 3D data as first step for a fall detection system. 

\noindent \textbf{Human Detection.}
One of the key challenges in human detection is occlusion, which increases the difficulty in human localization. Several methods \cite{DBLP:conf/iccv/TianLWT15} use part-based model to describe the pedestrian in occlusion handling, which learn a series of part detectors and design some mechanisms to fuse the part detection results to localize partially occluded pedestrians. Besides the part-based model, Zhou \etal \cite{DBLP:conf/iccv/ZhouY17} propose to jointly learn part detectors to exploit part correlations and reduce the computational cost. Wang \etal \cite{DBLP:conf/cvpr/WangXJSSS18} introduce a novel bounding box regression loss to detect pedestrians in the crowd scenes. Zhang \etal \cite{DBLP:conf/cvpr/Zhang0S18} propose to utilize channel-wise attention in convnets allowing the network to learn more representative features for different occlusion patterns in one coherent model. Zhang \etal \cite{DBLP:conf/eccv/ZhangWBLL18} design an aggregation loss to enforce proposals to be close and locate compactly to the corresponding objects. Zhou \etal \cite{DBLP:conf/eccv/ZhouY18} design a method to detect full body and visible part estimation simultaneously to further estimate occlusion.
Although numerous pedestrian detection methods are presented in literature, how to robustly detect each individual human in extremely crowded scenarios is still one of the most critical issues for human detectors.

\section{JointDet}

\subsection{Framework Overview}
\begin{figure}[t]
\centering
\subfigure[Statistical head-body ratio]{
\label{fig:ratio} 
\includegraphics[width=0.438\linewidth]{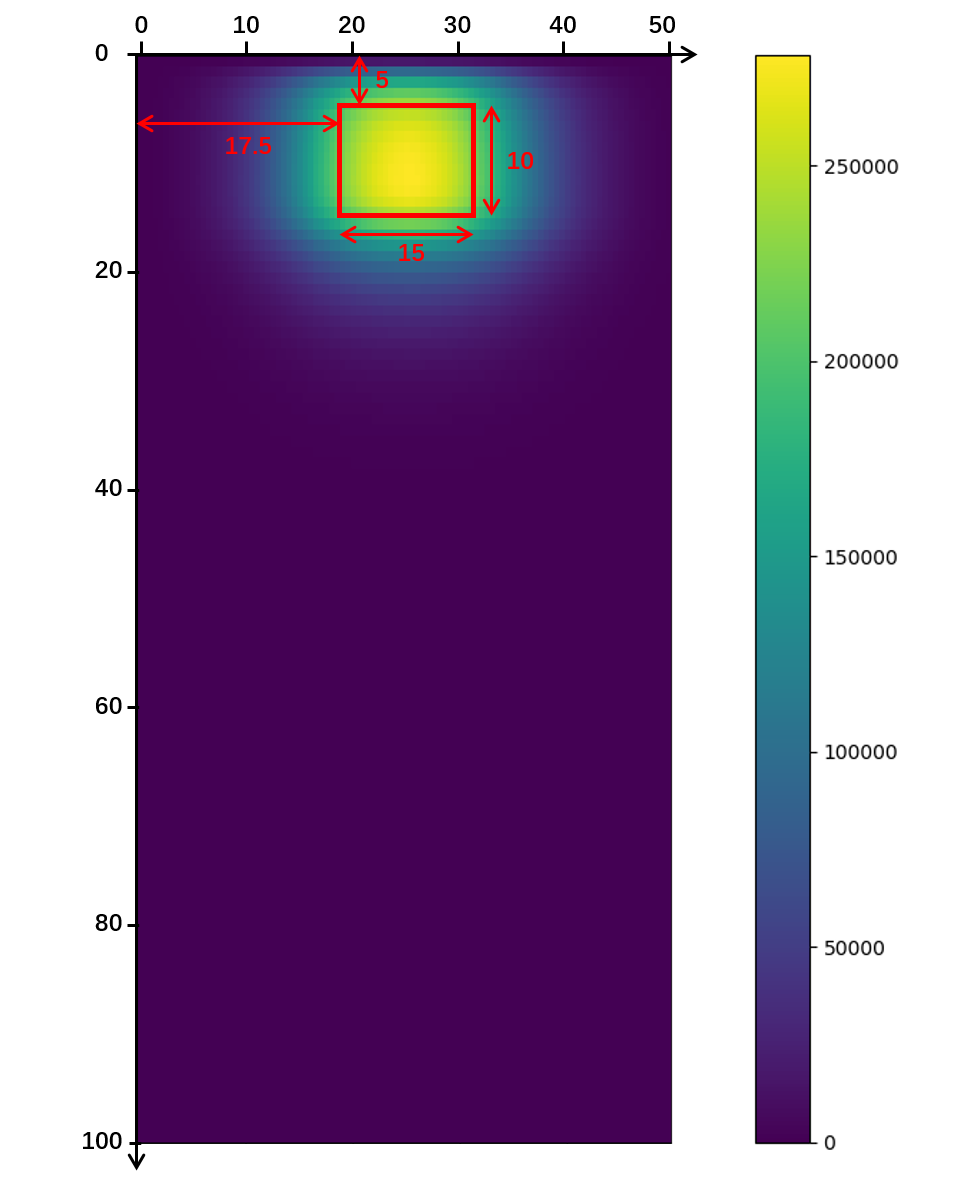}}
\ \ \ \ \ 
\subfigure[Examples]{
\label{fig:com-pro} 
\includegraphics[width=0.477\linewidth]{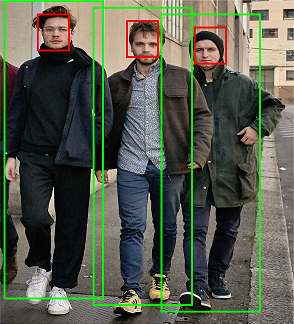}}
\caption{(a) The statistical head-body ratio to calculate the human body proposal based on the head proposal. (b) Red: head proposals; Green: inferred body proposals from head proposals via the statistical head-body ratio.}
\end{figure}

The overall framework is shown in Figure \ref{fig:framework}. We first utilize RPN to generate head proposals, then apply a statistical head-body ratio on these head proposals to obtain full-body proposals. The specific head-body ratio is shown in Figure \ref{fig:ratio} that is statistically obtained based on all human head-body pairs in the CrowdHuman dataset. After that, the head and full-body proposals are sent into two parallel R-CNN branches, respectively. Since the body proposals are obtained coarsely according to the head proposals, we adopt the cascade training strategy proposed by \cite{DBLP:journals/corr/abs-1712-00726} for the full-body R-CNN branch to regress more accurate results, where the full-body branch is passed through twice in the training and inference phases. The advantages of this framework can be summarized as follows:
\begin{itemize}
\setlength{\itemsep}{0pt}
\setlength{\parsep}{0pt}
\setlength{\parskip}{0pt}
\item \emph{A more efficient way to get head and human proposals.} The aspect ratio of head is almost fixed and we just need to tile anchors with one aspect ratio to obtain head proposals. In contrast, the human body has a wide range of aspect ratios because of its deformability and various postures. Tiling anchors to generate human proposals needs to preset a couple of aspect ratios that greatly reduce efficiency. To solve this issue, we use a statistical head-body ratio on head proposals to directly obtain human proposals for free.
\item \emph{Decoupling the classification task.} The two parallel R-CNNs only concentrate on detection of one class of object, \ie, head or human body. This design decouples two tasks into separate branches, which is beneficial to make targeted optimization respectively, \eg, using cascade strategy to improve the accuracy of calculated human proposals.
\end{itemize}

\subsection{Relationship Discriminating Module}
RDM is designed to learn to discriminate the relationships between the head-body pairs with larger Intersection over Head-box (IoH). The detail expression of IoH is as below:

\begin{equation}
\begin{aligned}
\mbox{IoH}=\frac{\mbox{Area of Overlap}}{\mbox{Area of Head-box}}.
\end{aligned}
\end{equation}

The structure of RDM is three stacked fully-connected layers, whose channel setting is same as the classification branch in R-CNN. The process of head-body pair matching and relationship prediction is described in Line $1$ to $9$ of Algorithm \ref{alg:rdm}. During training, when the matched pair belongs to one person, its ground-truth is $1$, otherwise $0$. We use the binary cross-entropy loss to optimize RDM. In addition, we set the batch size to be $512$, where the proportion of positive and negative examples is set as $1$:$3$.

In the inference phase, we gather the mismatched head detections that have low relationship score or are not matched through IoH, which is demonstrated in Line $10$ to $12$ of Algorithm \ref{alg:rdm}. There are two situations with the mismatched head detection: 1) The human body corresponding to this head is suppressed by NMS; 2) This head detection is a false positive without a corresponding body. Therefore, we use these mismatched heads and the body detections before NMS to perform matching and relationship discrimination again, as described in Line $14$ to $22$ of Algorithm \ref{alg:rdm}. If one mismatched head gets strong relationship response in the second time, then we recall the corresponding body detection (described in Line $23$ to $25$). In contrast, if one mismatched head fails again, then we treat it as a false positive, removing it from the results (described in Line $26$ to $28$).

\begin{algorithm}[!t]
\begin{small}
\caption{Relationship Discriminating Module}
\label{alg:rdm}
\begin{algorithmic}[1]
\Require{$ \mathcal{H},\ \mathcal{B}_{1},\ \mathcal{B}_{2},~\mathcal F,~\lambda,~\beta_1,~\beta_2$
\newline $\mathcal{H}$ is a set of head detections after NMS
\newline $\mathcal{B}_{1}$, $\mathcal{B}_{2}$ are a set of body detections before and after NMS
\newline $\mathcal{F}$ is the feature map of the P2 level in the FPN structure
\newline $\lambda$ is the IoH threshold while matching
\newline $\beta_1$, $\beta_2$ are the low and high relationship score thresholds
}
\Ensure{$ \mathcal{D}_{h},\ \mathcal{D}_{b}$
\newline $\mathcal{D}_{h}$, $\mathcal{D}_{b}$ are final head and body detections after post process
}
\vspace{3mm}
\newline \emph{/*- - - - - - -Find Mismatched Head for Post-process- - - - - -*/}
\State{$\mathcal{H}_m \leftarrow{} \varnothing $} ($\mathcal{H}_m$ is the set of mismatched head detections)
\For{$h_i \in \mathcal{H}$} 
	\State{$Score \leftarrow{} \varnothing $}
	\For{$b_j \in \mathcal{B}_2$}
		\If{$\mathrm{IoH(}h_{i},b_{j})>\lambda$}
    		\State{$Feat \leftarrow \mathrm{Concat(}{\mathrm{RoIPool}(\mathcal{F}, h_i),\mathrm{RoIPool}(\mathcal{F}, b_j))}$}
    		\State{$Score \leftarrow{Score\cup \{\mathrm{RDM(}Feat)\}}$}
		\EndIf
	\EndFor
	\If{$\mathrm{max(}Score) < \beta_1$ or $Score = \varnothing$}
		\State{$\mathcal{H}_m \leftarrow{\mathcal{H}_m \cup {h_i}}$}
	\EndIf
\EndFor
\vspace{3mm}
\newline \emph{/*- - - - - - - - - - - - -Post-process Method- - - - - - - - - - - - -*/}
\State{$\mathcal{D}_{h} \leftarrow{} \mathcal{H},\ \mathcal{D}_{b} \leftarrow{} \mathcal{B}_2$}
\For{$h_i \in \mathcal{H}_m$}
	\State{$Score \leftarrow{} \varnothing $}
	\For{$b_j \in \mathcal{B}_1$}
		\If{$\mathrm{IoH(}h_{i},b_{j})>\lambda$}
    		\State{$Feat \leftarrow \mathrm{Concat(}{\mathrm{RoIPool}(\mathcal{F}, h_i),\mathrm{RoIPool}(\mathcal{F}, b_j))}$}
    		\State{$Score \leftarrow{Score\cup \{\mathrm{RDM(}Feat)\}}$}
		\EndIf
	\EndFor
\vspace{1.5mm}
\newline
\vspace{0mm}
	\quad\ \emph{/*- - - - - - -Recall Suppressed Body Detections- - - - - -*/}
	\If{$\mathrm{max(}Score) > \beta_2$}
		\State{$\mathcal{D}_{b}\!\leftarrow{}\!\mathcal{D}_{b} \cup$ \{$b$ is the body detection with $\mathrm{max(}Score)$\}}
	\EndIf
\vspace{1.5mm}
\newline
\vspace{0mm}
	\quad\ \emph{/*- - - - - - - -Remove Head False Positives- - - - - - - - -*/}
	\If{$\mathrm{max(}Score) < \beta_1$ or $Score = \varnothing$}
		\State{$\mathcal{D}_{h} \leftarrow{} \mathcal{D}_{h} \setminus \{h_i\}$}
	\EndIf
\EndFor
\State{\Return{$\mathcal{D}_{h},\ \mathcal{D}_{b}$}}
\end{algorithmic}
\end{small}
\end{algorithm}

\subsection{Implementation Detail}

{\flushleft \textbf{Anchor Design.} }
At each location of the detection layer, we only associate one specific scale of anchors (\ie, $2S$, where $S$ represents the downsampling factor of the detection layer) and one aspect ratio (\ie, $1.25$). In total, there are $A=1$ anchors per level and they cover the scale range $8-128$ pixels across different levels with respect to the input image. 

{\flushleft \textbf{Sample Matching.} }
During training, anchors and proposals need to be divided into positive and negative samples. Specifically, samples are assigned to ground-truth boxes using an IoU threshold of $\theta_{p}$, and to background if their IoU is in $[0, \theta_{n})$. If an anchor is unassigned, which may happen with overlap in $[\theta_{n}, \theta_{p})$, it is ignored during the training phase. We set $\theta_{n}=0.3$ and $\theta_{p}=0.7$ for the RPN stage, and $\theta_{n}=0.5$ and $\theta_{p}=0.5$ for the R-CNN stage. 

{\flushleft \textbf{Loss Function.} }
The whole network is optimized by ${\cal L}={\cal L}_\text{RPN} + \lambda_1 {\cal L}_\text{Head} + \lambda_2 {\cal L}_\text{Body} + \lambda_3 {\cal L}_\text{RDM}$, where ${\cal L}_\text{RPN}$, ${\cal L}_\text{Head}$ represent the classification and regression loss of RPN and the head R-CNN branch, which are the same as those proposed in \cite{DBLP:journals/pami/RenHG017}. ${\cal L}_\text{Body}$ contains two-stage classification and regression loss as Cascade R-CNN \cite{DBLP:journals/corr/abs-1712-00726}. ${\cal L}_\text{RDM}$ is the log softmax loss over two classes, which indicates whether the head-body pair belong to one person. The loss weight coefficients ${\lambda_1}$, ${\lambda_2}$ and ${\lambda_3}$ are used to balance different loss terms and we empirically set them as $1$ in all the experiments.

{\flushleft \textbf{Initialization.} }
The backbone network is initialized by the ImageNet \cite{DBLP:journals/ijcv/RussakovskyDSKS15} pretrained ResNet-50. The parameters of newly added layers in RPN are initialized by the normal distribution method, and the parameters in R-CNN are initialized by the MSRA normal distribution method \cite{DBLP:conf/iccv/HeZRS15}.

{\flushleft \textbf{Optimization.} }
We fine-tune the model using SGD with $0.9$ momentum, $0.0001$ weight decay. The proposed JointDet is trained on $16$ GTX 1080Ti GPUs with a mini-batch $2$ per GPU for CrowdHuman and Caltech-USA, and the mini-batch size for Citypersons is $1$ per GPU. Each mini-batch involves $512$ RoIs per image. Multi-scale training and testing are not applied to ensure fair comparisons with previous methods. We implement JointDet using the PyTorch~\cite{paszke2017pytorch} library. The specific settings of training process for different datasets are described in next sections.

{\flushleft \textbf{Evaluation Metric.} }
Following \cite{DBLP:conf/cvpr/DollarWSP09}, the log-average miss rate over $9$ points ranging from $10^{-2}$ to $10^0$ FPPI (\ie, $\text{MR}^{-2}$) is used to evaluate the performance of the detectors. We report the detection performance for instances in head and full-body (\ie, human) categories.

\section{Experiments}
In this section, we perform extensive experiments on the CrowdHuman, CityPersons and Caltech-USA datasets to verify the effectiveness of the proposed framework.

\subsection{CrowdHuman Dataset}
CrowdHuman is a benchmark dataset to better evaluate detectors in crowd scenarios. It is large, rich-annotated, high-diversity and contains $15,000$, $4,370$ and $5,000$ images for training, validation and testing subsets, respectively. There are totally $470k$ human instances from the training and validation subsets, and $22.6$ persons per image, with various kinds of occlusions in the dataset. Each human instance is annotated with a head bounding-box, human visible-region bounding-box and human full-body bounding-box. The images and annotations of the training and validation subsets are made freely available to academic for scientific use, while only the images of the testing subset are released and the corresponding annotations are held-out. Since the online evaluation server is not available until now, all our models are trained on the CrowdHuman training subset and tested on the validation subset. During the training phase, the input images are resized so that their short edges are at $800$ pixels while the long edges should be no more than $1333$ pixels at the same time. We train JointDet with the initial learning rate $0.04$ for the first $16$ epochs, and decay it by $10$ and $100$ times for another $6$ and $3$ epochs.

\begin{table}[t]
\renewcommand\arraystretch{1.2}
\begin{center}
\caption{$\text{MR}^{-2}$ performance of different methods on CrowdHuman. Lower $\text{MR}^{-2}$ mean better performance.}
\label{tab:crowdhuman}
\setlength{\tabcolsep}{5.5pt}
\begin{tabular}{|c|c|c|c|}
\hline
Source & Method & Head & Human \\
\hline
\multirow{2}{*}{CrowdHuman}& FPN-Head & 52.1 & - \\
& FPN-Human & - & 50.4 \\
\hline
\multirow{5}{*}{Ours} & FPN-Head & 48.9 & - \\
& FPN-Human & - & 49.7 \\
& FPN-Human-Cascade & - & 49.2 \\
& JointDet w/o RDM & 48.7 & 47.0 \\
& JointDet & \bf{48.3} & \bf{46.5} \\
\hline
\end{tabular}
\end{center}
\end{table}

{\flushleft \textbf{Baseline.} }
Before delving into our proposed framework of joint head and human detection, we first build two strong baselines based on FPN \cite{DBLP:conf/cvpr/LinDGHHB17} for these two tasks, respectively. We set anchor scale to $2S$ in the head baseline and $8S$ in the full body baseline, where $S$ represents the stride size of each pyramid level. After considering the human body shape, we modify the height \emph{vs.} width ratios of anchors as $\{0.5$:$1$, $1$:$1$, $2$:$1\}$ for all the experiments related to human detection. While for head detection, the ratios are set to $1.25$:$1$. As shown in Table \ref{tab:crowdhuman}, the baseline of head detection, denoted as FPN-Head, achieves $48.9\%$ $\text{MR}^{-2}$ that is $3.2\%$ better than the head detection baseline in CrowdHuman (\ie, $52.1\%$). And the baseline of human detection, denoted as FPN-Human, obtains $49.7\%$ $\text{MR}^{-2}$ that is $0.7\%$ better than the full-body detection baseline in CrowdHuman (\ie, $50.4\%$). Thus, the detectors trained for the head and human respectively are two strong baselines to verify the effectiveness of our proposed framework.

\begin{figure*}[t]
\centering
\subfigure[Human]{
\label{fig:human}
\includegraphics[width=1.0\linewidth]{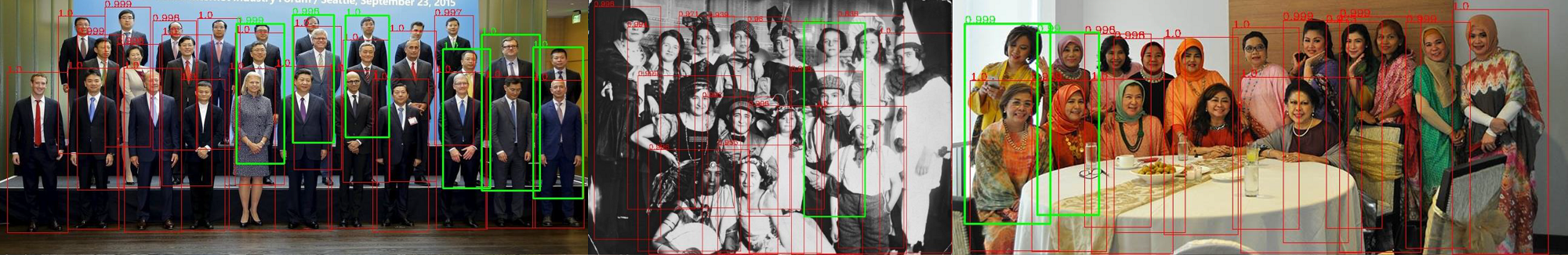}}
\subfigure[Head]{
\label{fig:head}
\includegraphics[width=1.0\linewidth]{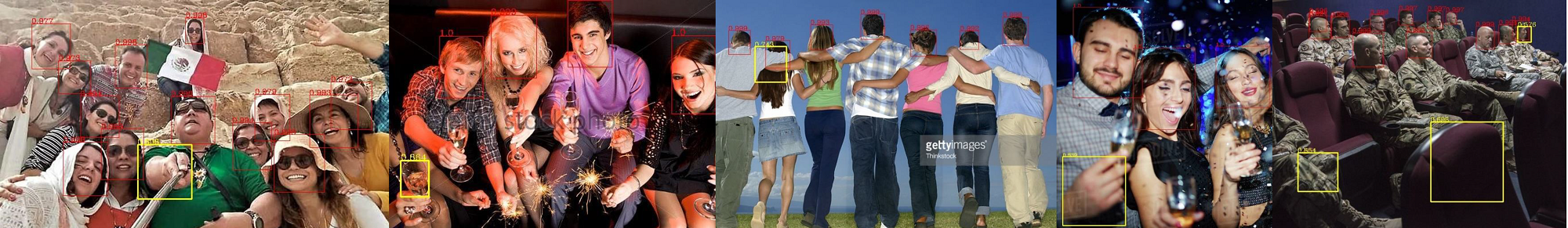}}
\caption{Qualitative results of JointDet on CrowdHuman. Red bounding boxes represent original results. Green bounding boxes represent recalled human results via RDM. Yellow bounding boxes represent removed head results via RDM.}
\label{fig:example}
\end{figure*}

{\flushleft \textbf{Ablation Study on Joint Detection.} }
As illustrated in Table \ref{tab:crowdhuman}, after jointing head and human detection in a single detection framework, we achieve $48.7\%$ $\text{MR}^{-2}$ for head detection and $47.0\%$ $\text{MR}^{-2}$ for human detection. Comparing to the baselines that each task is executed with a separate network, the proposed joint framework not only merges these two tasks into a single network so as to greatly improve the detection efficiency, but also has better $\text{MR}^{-2}$ performance, \ie, from $48.9\%$ to $48.7\%$ for head detection and from $49.7\%$ to $47.0\%$ for human detection. The $2.7\%$ improvement on human detection demonstrates the effectiveness of proposed proposal generation method. Notably, we use the cascade training strategy on the full-body R-CNN branch in our joint framework. To have a fair comparison, we train another human detection baseline, denoted as FPN-Human-Cascade, where the R-CNN branch is also passed through twice in the training and inference phases. FPN-Human-Cascade obtains $49.2\%$ $\text{MR}^{-2}$, which still has a large gap with the joint result of human detection. These results demonstrate the effectiveness of the joint framework of head and human detection.

{\flushleft \textbf{Ablation Study on RDM.} }
The final model of our proposed method is formed by adding the RDM on the joint framework of head and human detection. All the training and testing settings are consistent with previous experiments. The three hyperparameter is set as below: matching IoH threshold $\lambda$ is set to $0.7$, relationship score thresholds $\beta_1$ and $\beta_2$ are set to $0.1$ and $0.9$, respectively. As demonstrated in Table \ref{tab:crowdhuman}, after utilizing the head location information to recall the suppressed human bounding boxes, the $\text{MR}^{-2}$ of human detection is improved from $47.0\%$ to $46.5\%$. The advancement indicates that the proposed RDM does recall some human detections suppressed by NMS as shown in Figure \ref{fig:human}, making our JointDet robust to heavy occlusion in human detection. On the other hand, using the learned head-body relationship can also reduce some head false positives as shown in Figure \ref{fig:head}, boosting the $\text{MR}^{-2}$ of head detection from $48.7\%$ to $48.3\%$ and allowing our head detector to perform well in complex scenarios.

\subsection{CityPersons Dataset}
\begin{figure}[t]
\centering
\includegraphics[width=1.0\linewidth]{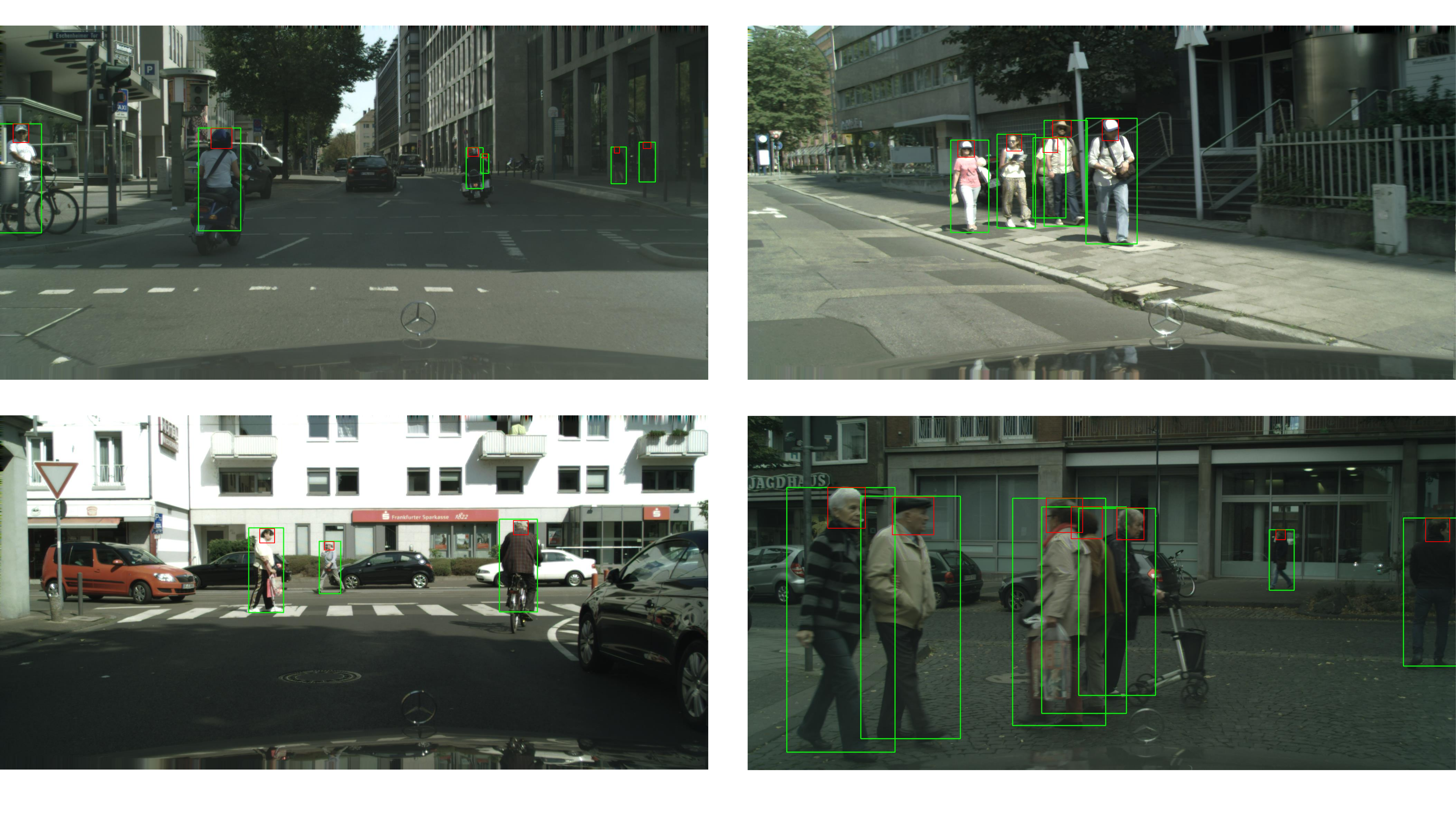}
\caption{Illustrations of additional head annotations of CityPersons dataset. Green: original pedestrian annotations. Red: additional head annotations.}
\label{fig:citypersons-head}
\end{figure}

CityPersons serves as a widely used benchmark dataset for pedestrian detection, which is built upon the semantic segmentation dataset Cityscapes \cite{DBLP:conf/cvpr/CordtsORREBFRS16}. It is recorded across $18$ different cities in Germany with $3$ different seasons and various weather conditions. The dataset includes $5,000$ images ($2,975$ for training, $500$ for validation, and $1,525$ for testing) with $\sim\negmedspace35,000$ manually annotated persons plus $\sim\negmedspace13,000$ ignore region annotations. Both the bounding boxes and visible parts of pedestrians are provided and there are approximately $7$ pedestrians in average per image. \textbf{For each annotated pedestrian instance, we additionally label the corresponding head bounding box}. The newly annotated head bounding box is within the scope of the original body bounding box. If the head is partly occluded, the annotators are asked to complete the invisible part. Some illustrations of additional head annotations are shown in Figure \ref{fig:citypersons-head}. The proposed JointDet detector is trained on the training set and evaluated on the validation set. Following the experiment settings in previous works \cite{DBLP:conf/cvpr/WangXJSSS18,DBLP:conf/eccv/ZhangWBLL18}, we enlarge input images by $1.3$ times. The initial learning rate is set to $0.02$ for the first $26$ epochs, and is decreased to $0.002$ and $0.0002$ for another $9$ and $5$ epochs, respectively.

We compare JointDet with TLL(MRF) \cite{Song_2018_ECCV}, Adapted FasterRCNN \cite{DBLP:conf/cvpr/ZhangBS17}, ALFNet \cite{Liu_2018_ECCV}, Repulsion Loss \cite{DBLP:conf/cvpr/WangXJSSS18}, PODE+RPN \cite{DBLP:conf/eccv/ZhouY18}, OR-CNN \cite{DBLP:conf/eccv/ZhangWBLL18} on the CityPersons validation subset in Table \ref{tab:cityperson-val}. Similar with previous works, we evaluate the final model on the Reasonable subset of the CityPersons dataset. The proposed method surpasses all published methods and reduces the $\text{MR}^{-2}$ of state-of-the-art results from $11.0\%$ to $10.23\%$ with $0.77\%$ improvement compared with the second best method \cite{DBLP:conf/eccv/ZhangWBLL18}, demonstrating the superiority of the proposed method in pedestrian detection.

\begin{table}[t]
\renewcommand\arraystretch{1.1}
\centering
\caption{$\text{MR}^{-2}$ performance on the CityPersons validation set. The scale indicates the enlarge number of original images in training and testing.}
\setlength{\tabcolsep}{5pt}
\begin{tabular}{|c|c|c|c|}
\hline
Method &Backbone &Scale &{\em Reasonable} \\ 
\hline
TLL(MRF) & ResNet-50 &- &14.40 \\
Adapted FasterRCNN & VGG-16 &$\times$1.3 &12.97 \\ 
ALFNet & VGG-16 &$\times$1 &12.00 \\
Repulsion Loss & ResNet-50 &$\times$1.3 &11.60 \\ 
PODE+RPN & VGG-16 &- &11.24 \\
OR-CNN & VGG-16 &$\times$1.3 &11.00 \\ 
\hline
JointDet (Ours) & ResNet-50 &\textbf{$\times$1.3} &\textbf{10.23} \\
\hline
\end{tabular}
\label{tab:cityperson-val}
\end{table}

\begin{figure}[t]
\centering
\includegraphics[width=1.0\linewidth]{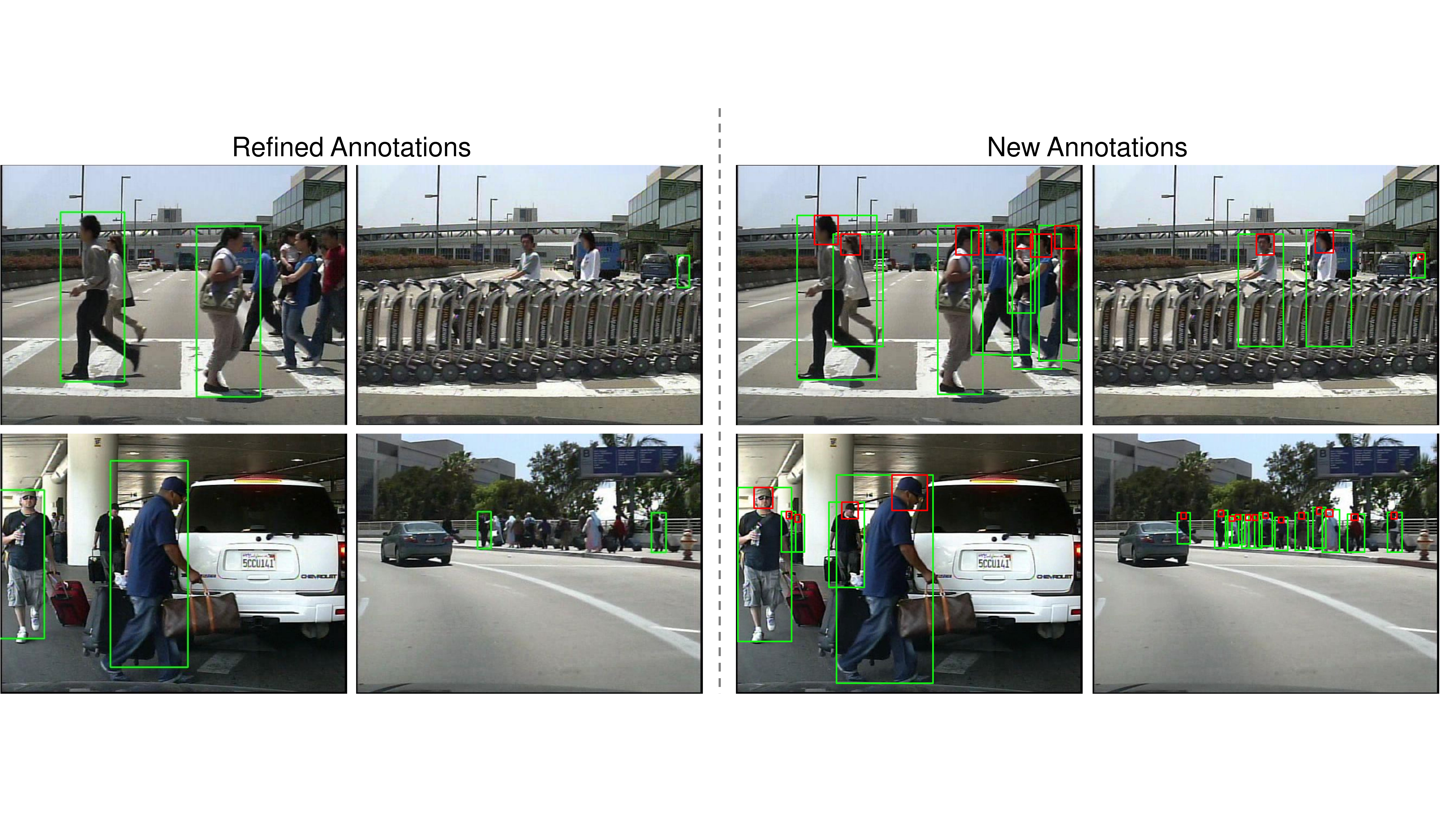}
\caption{Left: refined annotations provided by Zhang \etal Right: our new pedestrian body and head annotations. For simplicity, we do not draw the annotations of visible regions.}
\label{fig:caltech-new}
\end{figure}

\subsection{Caltech-USA Dataset}
Caltech-USA is one of the most popular and challenging datasets for pedestrian detection, which comes from approximately $10$ hours $30$Hz VGA video recorded by a car traversing the streets in the greater Los Angeles metropolitan area. The training and testing sets contain $42,782$ and $4,024$ frames, respectively. The commonly used $10\times$ training annotations \cite{DBLP:conf/cvpr/ZhangBOHS16} of Caltech-USA are refined automatically with only $16,376$ poor-quality instances in the training set. \textbf{We re-annotate the dataset manually, with a total of $32,273$ instances in the training set and $1,123$ instances in the testing set}. The labeling rule and method are consistent with the original ones \cite{DBLP:conf/cvpr/DollarWSP09}. \textbf{With the help of new pedestrian annotations, we also label their corresponding head bounding boxes}. Figure \ref{fig:caltech-new} shows the comparison of our new annotations and the widely used annotations provided by \cite{DBLP:conf/cvpr/ZhangBS15}. It is obvious that the quality of our new annotations is higher. The detailed analysis of the impact of our new annotations is described below and we use the new annotations to analyze the proposed JointDet method in next section. Following the experiment settings in \cite{DBLP:conf/cvpr/WangXJSSS18,DBLP:conf/eccv/ZhangWBLL18}, we train the proposed method using $2\times$ scale of the image size. The initial learning rate is $0.04$ for the first $4$ epochs, and is reduced by $10$ and $100$ times for another $2$ and $1$ epochs.

\begin{table}[h]
\renewcommand\arraystretch{1.2}
\centering
\caption{Effects of different training annotations on different validation annotations of the Caltech-USA dataset. We use the FPN baseline detector for clarification. `Refined Annotation' indicates the refined annotations by \cite{DBLP:conf/cvpr/ZhangBOHS16}. `New Annotation' indicates the new annotations re-annotated by us.}
\setlength{\tabcolsep}{3pt}
\begin{tabular}{|c|c|c|}
\hline
\diagbox{Training}{$\mbox{MR}^{-2}$}{Testing} & Refined Annotation & New Annotation\\
\hline
Refined annotation &4.31 &16.52 \\
New annotation &3.26 &14.26 \\
\hline
\end{tabular}
\label{tab:caltech-anno}
\end{table}

Here we first analyze the effect of our sanitised version of the annotations. As shown in Table \ref{tab:caltech-anno}, using the refined annotations provided by Zhang \etal~\cite{DBLP:conf/cvpr/ZhangBOHS16} for training, the FPN detector achieves $4.31\%$ $\text{MR}^{-2}$ on the refined testing set. It is reduced to $3.26\%$ with our re-annotated annotations as training set, indicating our new annotations possesses higher quality. When evaluating on our new testing annotations, performances of both detectors drop significantly, \ie, from $4.31\%$ to $16.52\%$ and from $3.26\%$ to $14.26\%$, which also verify the higher quality of our testing annotations. By statistics, our new annotations have a total of $32,273$ and $1,123$ ground truths in training and testing sets respectively, while the refined version in \cite{DBLP:conf/cvpr/ZhangBOHS16} only has $16,376$ and $912$ instances. Since the benchmark of the original annotations are reaching saturation, our new annotations can serve as a new evaluation metric.

\begin{figure}[t]
\centering
\includegraphics[width=0.95\linewidth]{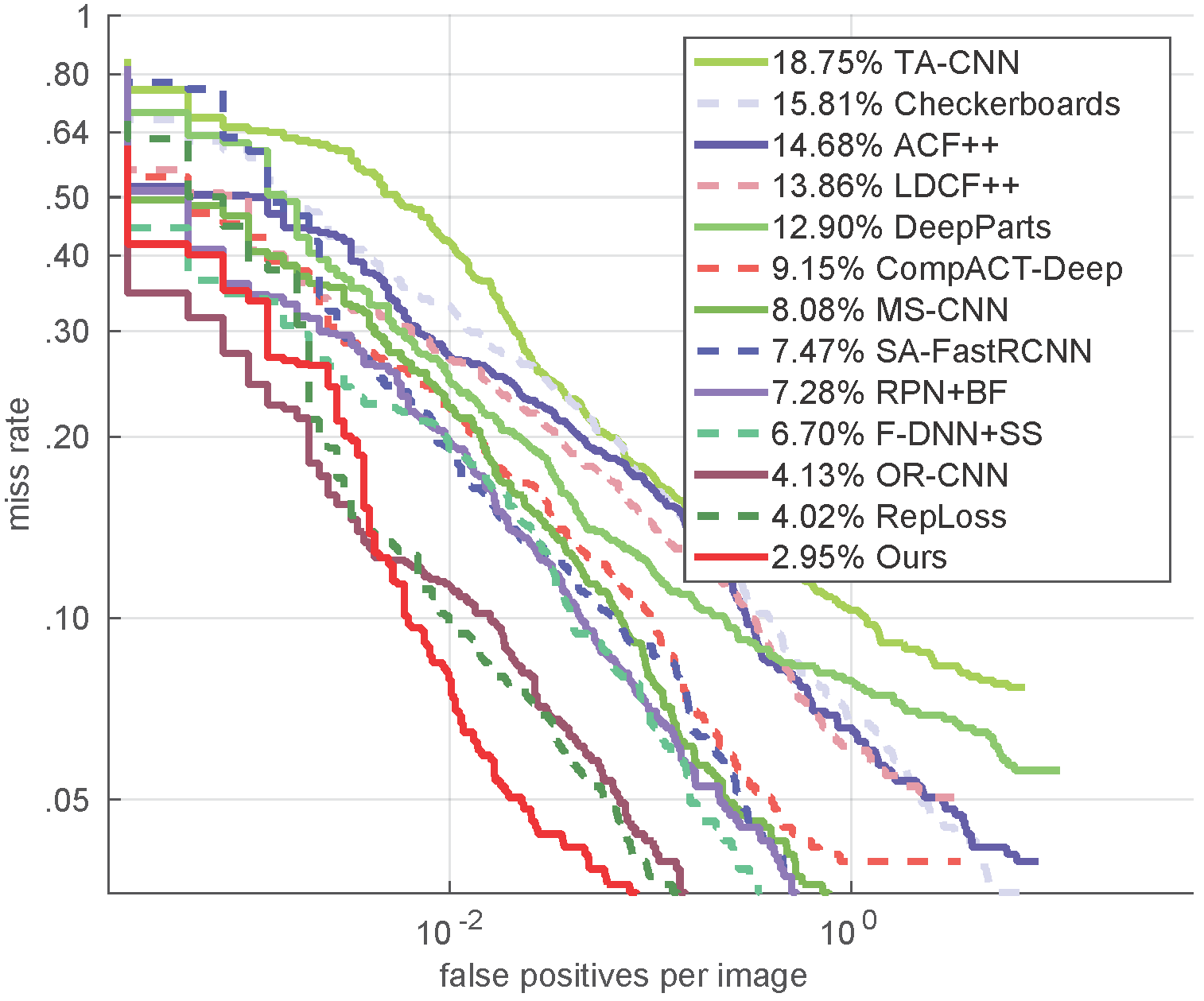}
\caption{Comparisons with the state-of-the-art methods on the Caltech-USA dataset. The scores in the legend are the ${\text{MR}}^{-2}$ scores of the corresponding methods.}
\label{fig:caltech}
\end{figure}

Figure \ref{fig:caltech} shows the comparison of the JointDet method with other state-of-the-art methods \cite{DBLP:conf/eccv/CaiFFV16,DBLP:conf/iccv/CaiSV15,DBLP:conf/cvpr/CosteaN16,DBLP:conf/wacv/DuELD17,DBLP:journals/tmm/LiLSXFY18,DBLP:conf/cvpr/MaoXJC17,DBLP:conf/icpr/Ohn-BarT16a,DBLP:conf/iccv/TianLWT15,DBLP:conf/cvpr/WangXJSSS18,DBLP:conf/eccv/ZhangLLH16,DBLP:conf/cvpr/ZhangBS15} on the Caltech-USA refined testing set. All the reported results are evaluated on the widely-used Reasonable subset, which only contains pedestrians with at least $50$ pixels tall and occlusion ratio less than $35\%$. The proposed method outperforms all other methods by producing $2.95\%$ $\text{MR}^{-2}$.

\subsection{Discussion}

{\flushleft \textbf{Head and Human Annotation.} } The proposed method requires both head and human annotations, which is feasible in practical applications and academic research with the consideration of the following two aspects: 1) If the human or the heads are annotated, another kind of annotations is easy to obtain via the automatic labelling method (\eg, using a trained head or human detector) or the semi-automatic labelling method (\eg, manual correcting after pre-labelling); 2) We will release all new annotations of head and human to facilitate further studies of head and human detection.

{\flushleft \textbf{Occluded Head.} } 
The proposed method generates the human proposals based on the corresponding head proposals. The results of FPN-Human-Cascade ($49.2\%$) and JointDet w/o RDM ($47.0\%$) in Table \ref{tab:crowdhuman} have verify that this way of generating human proposals is better than using the RPN proposals. If the head is occluded, it maybe cause some human miss detection but has a slight impact due to: 1) Human with only head occluded is small number case, while occluded body is more common; 2) With help of human body context, RPN can generate proposals for some occluded heads. Thus, the occluded head has ignorable impact and our state-of-the-art human detection performance also confirms the above statement.

\section{Conclusion}
In this paper, we have presented a novel joint detection network to detect head and human simultaneously, which utilizes the learned relationship between heads and human bodies to recall the suppressed human detections and reduce head false positives. To sufficiently verify the effectiveness of these proposed components, we have made some efforts in the dataset: 1) providing a better version of Caltech-USA annotations with full body and visible region; 2) annotating the head bounding boxes of CityPersons and Caltech-USA. Consequently, the proposed JointDet detector achieves state-of-the-art performance on CrowdHuman, CityPersons and Caltech-USA. All new annotations, source codes and trained models will be public to facilitate further studies of head and human detection.

\clearpage

\bibliographystyle{aaai}
\bibliography{reference}
\end{document}